\documentclass[10pt,twocolumn,letterpaper]{article}

 \usepackage[pagenumbers]{cvpr} 

\usepackage{placeins}

\usepackage[dvipsnames]{xcolor}

\definecolor{cvprblue}{rgb}{0.21,0.49,0.74}
\usepackage[pagebackref,breaklinks,colorlinks,citecolor=cvprblue]{hyperref}

\def\paperID{5027} 
\def\confName{CVPR}
\def\confYear{2024}

\usepackage{multirow}
\usepackage{multicol}

\title{SurfaceAug: Closing the Gap in Multimodal Ground Truth Sampling}

\author{Ryan Rubel \and Nathan Clark\\
Noblis\\
2002 Edmund Halley Dr\\
{\tt\small \{ryan.rubel, nathan.clark, andrew.dudash\}@noblis.org}
\and Andrew Dudash
}

\begin{document}
\maketitle
\begin{abstract}
Despite recent advances in both model architectures and data augmentation, multimodal object detectors still barely outperform their LiDAR-only counterparts. This shortcoming has been attributed to a lack of sufficiently powerful multimodal data augmentation. To address this, we present SurfaceAug, a novel ground truth sampling algorithm. SurfaceAug pastes objects by resampling both images and point clouds, enabling object-level transformations in both modalities. We evaluate our algorithm by training a multimodal detector on KITTI and compare its performance to previous works. We show experimentally that SurfaceAug outperforms existing methods on car detection tasks and establishes a new state of the art for multimodal ground truth sampling.
\end{abstract}
\FloatBarrier
\section{Introduction}

The complementary characteristics of LiDARs and cameras make them an obvious candidate for sensor fusion. LiDAR point clouds are sparse and 3D, while images are dense and 2D. Intuitively, this combination is ideal for 3D object detection---the LiDAR excels at localization but struggles with recognition, while the camera excels at recognition but struggles with localization.

Yet rankings on benchmarks such as KITTI \cite{KITTI_benchmark} do little to support this intuition. At the time of writing, the best-performing 3D pedestrian detector uses only LiDAR \cite{casa2022}. The model \cite{VirConv} with the best performance on KITTI's 3D car detection task is a multimodal algorithm, but it outperforms the best LiDAR-only method \cite{casa2022} by a mere 3.16 AP.\footnote{As of Oct. 2, 2023, VirConv \cite{VirConv} is the top-ranked detector for KITTI's 3D car detection task at the moderate level. It outperforms CasA \cite{casa2022} with an AP of 87.2 vs. 84.04.} Why is this?




Recent studies \cite{MoCa, Context-guided} attribute this apparent underperformance of multimodal methods to a lack of powerful data augmentation techniques. Due to the inherent difficulty of preserving correspondences between 3D and 2D data, the augmentation strategies used in published multimodal detectors are generally weaker than those used by LiDAR-only methods. For instance, when the authors of VoxelNet \cite{VoxelNet} extended their method to incorporate image data in a subsequent paper, they wrote, ``...since we use both images and point clouds, some of the augmentation strategies used in the original work are not applicable to the proposed multimodal framework, e.g., global point cloud rotation.'' \cite{MVX-Net} While this issue was later partially resolved with the introduction of the multi-modality transformation flow \cite{MoCa}, object-level augmentations such as the rotations used in the original VoxelNet paper remained elusive. In this paper, we finally address this issue of multimodal object-level augmentations.




\begin{figure}[t!]
  \centering
  \includegraphics[width=1.067\columnwidth]{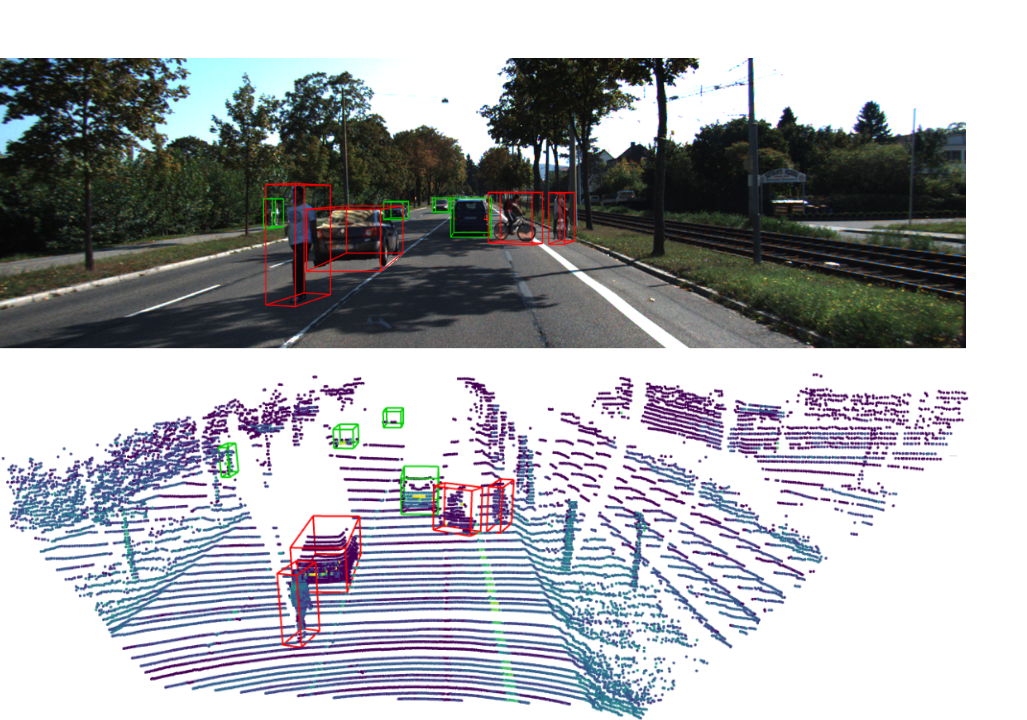}

   \caption{SurfaceAug pastes objects by resampling both the image and point cloud. Pictured: pasted objects are boxed in red; original objects are boxed in green.}
   \label{fig:intro}
\end{figure}

\FloatBarrier

\subsection{Contributions}
We propose a novel multimodal ground truth sampling algorithm that allows for limited similarity transformations of pasted objects in both point clouds and images. Our specific contributions are as follows:
\begin{itemize}
    \item We introduce a novel method of preserving multimodal consistency across out-of-plane transformations by exploiting an intermediate surface mesh representation.
    \item We formulate a set of ad hoc rules to ensure realistic placement and appearance of pasted objects.
    \item We propose an explainable, purely mathematical approach to LiDAR intensity simulation.
    \item We demonstrate the efficacy of our proposed algorithm by training and evaluating MVX-Net \cite{MVX-Net} on the KITTI dataset \cite{KITTI_raw}.
    \item We evaluate our method against existing ground truth sampling approaches and outperform them on car detection tasks.
\end{itemize}



\section{Related Work}
\label{sec:formatting}

\subsubsection{Object detection with LiDAR and camera} 



LiDAR-camera fusion algorithms are broadly categorized into early, deep, and late fusion paradigms \cite{FusionLevels}. Early fusion methods combine raw image and point cloud data prior to feature extraction. Early fusion is uncommon in the context of object detection,\footnote{Sindagi et. al \cite{MVX-Net} describe their PointFusion method as ``early fusion.'' However, PointFusion combines point data with extracted image features, rather than raw images. Image feature extraction lowers resolution and enlarges receptive field, lessening the impact of alignment errors.} as fusing raw data is particularly susceptible to input misalignment from point noise and calibration error. Deep fusion algorithms extract image and point cloud features independently before fusing them and predicting bounding boxes. Late fusion algorithms combine object proposals generated from separate LiDAR and camera pipelines. Provided that their component single-modality detectors have sufficiently high recall, late fusion methods are able to effectively filter out false positives and predict accurate bounding boxes by fusing and refining input proposals \cite{CLOCs, Fast-CLOCs}.

Most multimodal object detectors fall within the deep fusion paradigm. MVX-Net \cite{MVX-Net} combines image features from a pre-trained Faster-RCNN \cite{Faster-RCNN} with point data prior to voxel feature extraction \cite{VoxelNet}. Frustum PointNets \cite{Frustum_PointNets} trains an image-based detector to produce 2D bounding box proposals and uses a PointNet \cite{PointNet} to segment points within the frusta defined by proposed 2D boxes. A regression PointNet is trained to predict 3D bounding boxes from the segmented points. MV3D \cite{MV3D} projects LiDAR point clouds into bird's eye view (BEV) and front view pseudo-images and uses CNNs to extract features from the image and both pseudo-images. Features from each view are pooled based on region proposals prior to fusion and bounding box prediction.


\subsubsection{Ground truth sampling}
Ground truth sampling is a data augmentation strategy that synthetically adds objects of interest and their detection labels to existing data. This both helps combat class imbalance and promotes robustness against changes in setting. Echoing the widespread success \cite{Pattern-aware, WYSIWYG, SECOND} of ground truth sampling in improving LiDAR-based object detectors, recent works \cite{MoCa, Context-guided} have extended ground truth sampling to handle complementary image data. However, they neglect object-level transformations such as uniform scaling, translation \cite{Frustum_PointNets}, and rotation \cite{VoxelNet}, which are commonly used for data augmentation in training pipelines.



\subsection{Modeling LiDAR peculiarities}

\subsubsection{Missing points}

Due to stochastic environmental factors, some LiDAR beams fail to return to the sensor. Removing points to properly model occlusion of pasted objects proved beneficial to object detection performance \cite{WYSIWYG}, underscoring the importance of accurately modeling the native data distribution. Subsequent research showed minor gains from adjusting ground truth sampling to account for reduced LiDAR point density at greater distances \cite{Pattern-aware}. However, their correction merely downsampled pasted object point clouds, limiting its representation power. Recently, LiDARsim \cite{LiDARsim} narrowed the sim2real gap by training a neural network to drop certain points from a raycasted point cloud, emulating the failure of some LiDAR beams to return to the sensor. 

\subsubsection{Intensity}
LiDAR sensors provide an intensity field that is a function \cite{LIDAR-math-paper} of the optical power returned to the receiver. Intensity is frequently used as a feature in LiDAR-based object detection \cite{VoxelNet, MVX-Net, SECOND, PointPillars} because it adds semantic information to an otherwise purely geometric datatype. This has motivated attempts to simulate LiDAR intensity for realistic point cloud synthesis. Recent work trained a neural network on RGB-D and LiDAR point clouds to predict intensity values \cite{Learning-to-Predict-Lidar-Intensities}. LiDARsim \cite{LiDARsim} took a simpler approach, directly sampling intensity values from a mesh painted with the original intensities and thereby neglecting geometric influences \cite{LIDAR-math-paper}.



\begin{figure*}[t]
  \centering
  \includegraphics[width=1\linewidth]{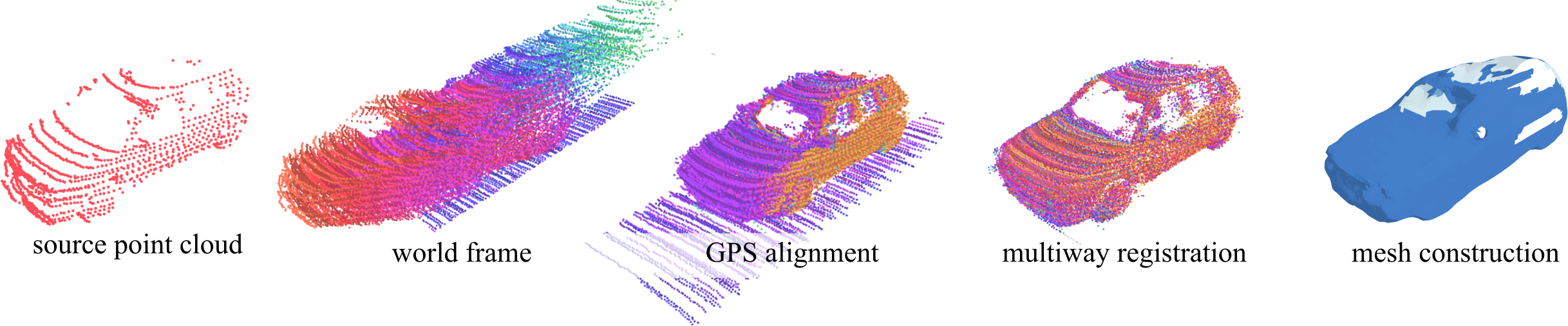}

   \caption{Cut object construction pipeline. We combine the ego car's localization estimates with sparse bounding box labels to transform LiDAR points into the world frame. After multiway registration via pose graph optimization, we construct a high-quality object mesh. Triangles in the mesh that are shown as transparent allow LiDAR rays to pass through, but not camera rays.}
   \label{fig:objconstruction}
\end{figure*}

\section{SurfaceAug}

In preprocessing, we construct a database of pastable ground truth objects, which we refer to as ``cut objects.'' We construct a surface mesh for each cut object.

During training, we randomly select cut objects from the database according to a predefined class sampling distribution. For each selected cut object, we sample a specially tailored similarity transformation---which we call the ``pasting transformation''---to paste its mesh into the scene. The pasted mesh is used to intercept simulated LiDAR and camera rays, which are in turn used to resample the point cloud and image, respectively. Objects are pasted in descending order of depth; the image and point cloud are resampled for each object sequentially. Finally, we generate a set of ground truth labels to match the sampled cut objects and their pasting transformations.

\subsection{Object cutting}
Prior to training, we use the KITTI dataset \cite{KITTI_raw, KITTI_benchmark} to construct a database of cut objects. Each cut object consists of a surface mesh, the object's cropped image, its instance mask (which we select from either \cite{KINS} or \cite{MonoCInIS}), its class and dimensions, an interpolant \cite{SciPy, Qhull} for sampling LiDAR intensities, and precomputed factors used in determining valid pose proposals. To avoid undesirable reconstruction artifacts, we only consider unoccluded objects for inclusion in the ground truth database.


\subsubsection{Point cloud accumulation}
Accurate image resampling relies on accurate resampling maps, which in turn rely on accurate mesh geometry. To improve the quality of constructed object meshes, we accumulate dense object point clouds by agglomerating LiDAR scans from KITTI \cite{KITTI_raw} drive sequences. However, KITTI poses several challenges:


\begin{itemize}
\item Only a fraction of the scenes are labeled.
\item Labels are provided in the ego car frame, but some objects move independently of the ego car.
\item Tracklets (which track the same labeled object across labeled scenes) are only provided for certain drive sequences.
\end{itemize}

We first transform the point clouds and labels into a static world frame. To get the LiDAR's pose as a function of time, we follow KITTI's approach of converting GPS+IMU data to poses, then interpolate \cite{PoseInterpolation} between these poses based on the provided LiDAR timestamps.

We create our own tracklets by matching the world-frame labels with the Hungarian algorithm \cite{HungarianAlgorithm} based on a heuristic combining geodesic distance and class and size agreement (cars are rigid and maintain constant dimensions over time).

The accumulated world-frame labels and point clouds allow us to aggregate LiDAR points that fall between consecutive object labels. We roughly estimate the world poses of objects in unlabeled frames by using LiDAR timestamps to interpolate \cite{PoseInterpolation} between the poses of their previous and next world frame labels. We use these estimated poses to collect object point clouds into a common object frame.

At this point, we have a coarsely aligned sequence of object point clouds in the object frame. For fine alignment, we perform multiway point cloud registration. To estimate the transformations between point clouds, we follow LiDARsim's approach \cite{LiDARsim}, using colored ICP \cite{Colored_ICP} with intensity features as colors. For global consistency, we construct and optimize a pose graph \cite{PoseGraphOptimization} for joint registration of object point clouds, jointly optimizing corrections to the coarse estimates. Finally, after registration, we remove outliers and use RANSAC \cite{RANSAC} to remove ground plane points.

Our registration approach only works for rigid objects. For deformable objects such as pedestrians and cyclists, we construct each mesh from a single LiDAR scan.

\subsubsection{Surface reconstruction}
We simulate object color by painting the object's source image onto its pasted mesh. We are simulating watertight objects, so the mesh's projection cannot have holes. Object meshes are constructed from accumulated LiDAR point clouds, but LiDAR typically penetrates glass, so car windows lack supporting LiDAR points. To ensure the mesh has no interior holes despite the missing window points, we employ screened Poisson reconstruction \cite{Screened-Poisson-Recon}. We ensure that LiDAR points still pass through windows by treating triangles with sparse support as invisible to LiDAR for the purpose of mesh intersection. Finally, we trim excess surface hallucinations extending beyond the object's bounding box label.

\subsection{Object pasting}

\begin{figure*}[t]
  \centering
   \includegraphics[width=1\linewidth]{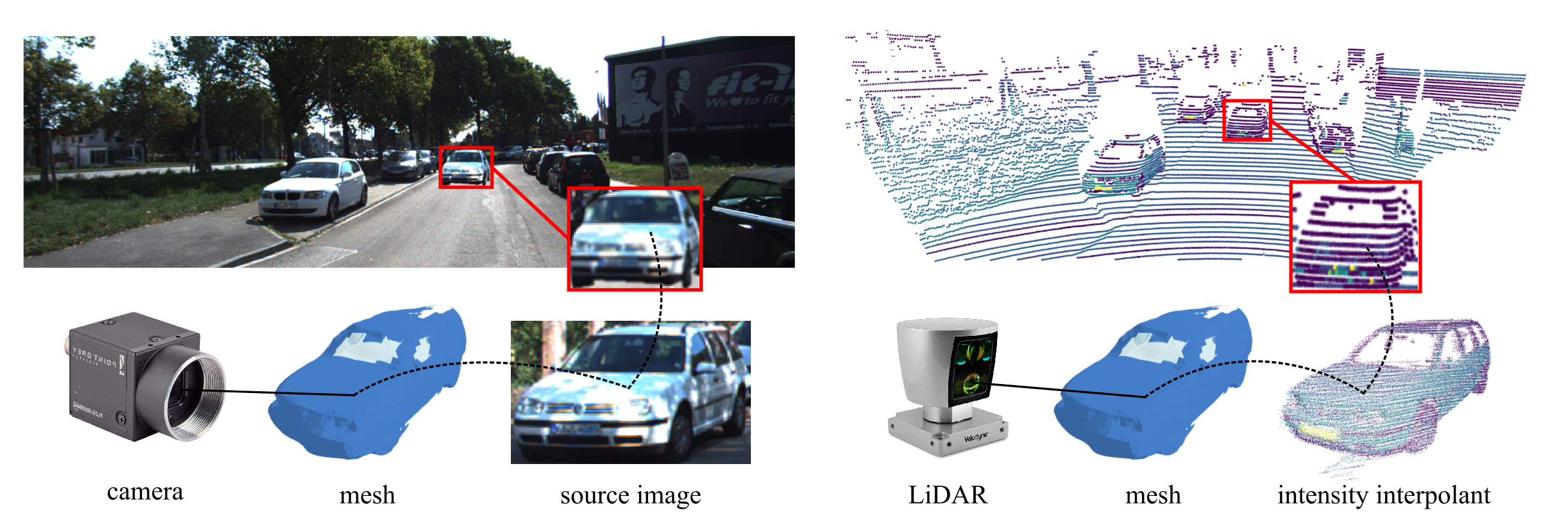}

   \caption{The SurfaceAug pipeline. Left: simulated camera rays are intersected with the pasted mesh, then transformed onto the source image. Colors are sampled from the source image and painted onto the destination image. Right: simulated LiDAR rays are intersected with the pasted mesh. Each intersection is transformed into the object frame to resample its intensity. Existing points occluded by the pasted mesh are removed.}
   \label{fig:surfaceaug}
\end{figure*}

\subsubsection{Image resampling}
We render a pasted object in a new pose by transforming the object's mesh to that pose and intersecting \cite{Moller-Trumbore} a virtual camera ray against it for each pixel. This associates each pixel with a 3D point on the surface of the object. We then transform these 3D points back to the cut object's pose in the source image and project them onto the image plane via the camera projection matrix $P$. Finally, we sample the source image at these points via bilinear interpolation and replace the corresponding pixels in the destination image.


\subsubsection{Anti-aliasing} Previous works \cite{CutPasteLearn, Context-guided} have shown that pasted images leave undesirable boundary artifacts, which can hinder generalization. To prevent this, we randomly apply a slight Gaussian blur to some pasted object images. To further guarantee a smooth transition from the pasted image to the original, we supersample pasted images by sampling each pixel with four rays rather than one. When sampling from the source image, we also sample its corresponding instance segmentation mask, yielding smooth blending weights around the edges.


\subsubsection{Simulating LiDAR geometry}
Our approach to pasting a cut object into a LiDAR point cloud consists of two steps. First, we remove the points from the original cloud that are occluded \cite{Moller-Trumbore} by the mesh. Second, we simulate new LiDAR points that intersect \cite{Moller-Trumbore} with the pasted mesh.

\newcommand{\ith}{^{(\text{i})}}
To simulate new LiDAR points, we use the LiDAR calibration parameters \cite{KITTI-calib-file} provided with the KITTI dataset to compute the outgoing ray direction $\Vec{r}\ith$ and origin $\Vec{o}\ith$ for laser $l\ith$ as a function of the rotation angle $\phi$ of the LiDAR assembly:



\begin{gather}
\begin{split}
    \Vec{r}\ith(\phi) =
    \begin{bmatrix}
    d_{xy}\ith \cos{\beta\ith}\\
    d_{xy}\ith \sin{\beta\ith}\\
    d \sin{\theta\ith}
    \end{bmatrix};
    & \hspace{3pt}
    \Vec{o}\ith(\phi) =
    \begin{bmatrix}
    -h_o\ith \cos{\beta\ith}\\
    h_o\ith \sin{\beta\ith}\\
    v_o\ith \cos{\theta\ith}
    \end{bmatrix}
\end{split}\\
    \begin{split}
    \beta\ith(\phi) = \phi - \alpha\ith \hspace{20pt} & d_{xy}\ith = d \cos{\theta\ith} - v_o\ith \sin{\theta\ith}
    \end{split}
\end{gather}

where $d>0$ is the (arbitrary) length of the LiDAR ray. The remaining terms are the intrinsic parameters \cite{LiDAR-calibration} for laser $l\ith$: $\alpha\ith$ (resp. $\theta\ith$) is the rotational (resp. vertical) correction factor and $h_o\ith$ (resp. $v_o\ith$) is the horizontal (resp. vertical) offset.



We simulate the LiDAR according to the firing pattern of a Velodyne HDL-64E S2 \footnote{Although KITTI \cite{KITTI_vision_meets_robotics, KITTI_are_we_ready} states that the point clouds are from a Velodyne HDL-64E \cite{HDL64E}, the point cloud patterns indicate that they are from an S2.} \cite{HDL64ES2}---that is, we fire the top and bottom blocks of lasers simultaneously. Of the simulated LiDAR rays, we keep those that intersect with the pasted mesh and whose intersections project onto the object's instance segmentation mask.





\subsubsection{Simulating LiDAR intensity}


LiDAR intensity is a function of many factors that cannot be readily determined from an object detection dataset like KITTI \cite{KITTI_raw, KITTI_benchmark} and is therefore infeasible to model directly. While this has motivated some to resort to learned methods \cite{Learning-to-Predict-Lidar-Intensities}, we prefer a more direct and explainable solution.

We model a function $f(\cdot)$ that, given a reference pose $H$ from which the LiDAR observes the object, maps a point $x$ on the object's surface to its intensity $f(x ; H)$. We begin with the following equation, which is trivially true:
\begin{equation}
    f(x ; H) = f(x ; H_0)\frac{f(x ; H)}{f(x ; H_0)}
    \label{eq:frac}
\end{equation}

for any other reference pose $H_0$. We assume that intensity is roughly proportional to the optical power $P$ received by the LiDAR sensor:

\begin{equation}
    f(x ; H) \approx kP(x ; H) \implies \frac{f(x ; H)}{f(x ; H_0)} \approx \frac{P(x ; H)}{P(x ; H_0)}
    \label{eq:proportional}
\end{equation}

We also assume that object surfaces are Lambertian and intercept the full LiDAR beam, allowing the application of a simplified form of the LiDAR range equation \cite{LIDAR-math-paper}:

\begin{equation}
    P(x ; H) = \frac{P_t D^2_r \eta_{\textit{atm}} \eta_{\textit{sys}} \rho(x)}{4R(x ; H)^2}\cos{{\alpha_{x ; H}}}
    \label{eq:lidar}
\end{equation}

where $P_t$ is the optical power transmitted by the LiDAR, $D_r$ is the receiver aperture diameter, $\eta_{\textit{atm}}$ is the atmospheric transmission factor, and $\eta_{\textit{sys}}$ is the system transmission factor (all assumed constant). $\rho(x)$ is the reflectance of the object's surface at $x$. For a given reference pose $H$, $R(x ; H)$ is the distance from the LiDAR sensor to the point $x$ and $\alpha_{x ; H}$ is the angle of incidence of the LiDAR ray with the object's surface at the point $x$.

We assume that $\rho(\cdot)$ is approximately constant on small scales:

\begin{equation}
    x \approx x_0 \implies \rho(x) \approx \rho(x_0)
    \label{eq:approx_constant}
\end{equation}

Combining \eqref{eq:frac}, \eqref{eq:proportional}, \eqref{eq:lidar}, and \eqref{eq:approx_constant} yields:
\begin{equation}
    \begin{split}
    f(x ; H) \approx f(x ; H_0) \frac{P(x ; H)}{P(x ; H_0)} \approx
    \frac{\cos{\alpha_{x ; H}}}{R(x ; H)^2} \cdot b \\
b = \frac{f(x ; H_0)R(x ; H_0)^2}{\cos{\alpha_{x ; H_0}}}
    \end{split}
\end{equation}

Since $x$ may not be among the original object points directly observed by the LiDAR, we model $b$ by building a linear interpolant \cite{SciPy, Qhull} $\text{lerp}(x ; H_0)$ over the original LiDAR points used to construct the cut object. We use the normals of the constructed mesh to estimate the angle of incidence $\alpha_{x ; H_0}$. Finally, we have:

\begin{equation}
    f(x ; H) \approx \frac{\cos{\alpha_{x ; H}}}{R(x ; H)^2} \cdot \text{lerp}(x ; H_0)
\end{equation}

We compute these values (including the interpolant) in log space, improving numerical stability and replacing the smooth arithmetic mean with a sharper geometric mean.

\subsection{Sampling valid pasting transformations}


To promote pose-invariant detection, we desire a diverse dataset full of objects observed with a wide variety of poses. However, most poses are not viable for cut object placement. We use rejection sampling to avoid any transformation that would (1) place an object in an unrealistic location, (2) cause the object to be occluded by the original scene, or (3) overly distort the resampled image of the object. We also incorporate slight random uniform scaling to improve scale invariance.



\subsubsection{Maintaining original viewpoint}

Because we construct our cut objects from the KITTI training set, we are only able to resample data from the portions of the cut object that were observed by the sensors.
When a cut object is pasted, it cannot expose faces that were concealed in the data used to construct it. To enforce this rule, for each cut object, we record which sides (front/back and left/right) faced the camera in the source image.

At augmentation time, any pose proposal that would reveal an object's unobserved front or back side is rotated 180º about its yaw axis. Then, if necessary, we add a reflection to keep the object's unobserved left or right side hidden. The combination of these two transformations ensures that every sampled transformation causes the pasted mesh to face toward the LiDAR and camera.






%

\subsubsection{Rejecting unrealistic and nonviable placements}

We avoid foreground occlusion as properly modeling such an occlusion in the image would require detailed knowledge of the occluding geometry, which may be unavailable.

All prospective object poses are projected onto the ground based on the average height of the points under the potential bounding box. This approach accounts for uneven terrain better than a precomputed ground plane estimate. After the potential pose has been translated to the ground, we reject poses behind existing LiDAR points, which suggest foreground occlusion. Prospective poses with too few ground points are likely occluded as well, so we reject them. Poses with large variance in ground point height are likely inside of an object or not on a level surface, so we reject them too. Finally, we reject any pasting transformations that would place an object behind an existing bounding box, as well as those that would intersect any of them.

\subsubsection{Limiting distortion}

Pasting a cut object at too extreme of an angle may require one face of the cut object to be upsampled heavily. Upsampling introduces quantization artifacts and exacerbates distortion from errors in the resampling maps. Therefore, we prevent the ratio of the pasted object's projected bounding box side lengths to those of the original from exceeding a predefined threshold $d_{max}$, which we empirically set to 1.5.

We record the lengths (in pixels) of the length and width of the cut object's projected bounding box label along the $u$ axis in the source image. At augmentation time, we recompute these projected lengths based on each pasting transformation. We reject those pasting transformations that would cause this ratio of projected lengths to exceed $d_{max}$.










\begin{table*}[t]
  \centering

  \begin{tabular}{@{}p{0.12\linewidth}|cccc|cccc|cccc@{}}

  &
  \multicolumn{4}{c|}{Car}&
  \multicolumn{4}{|c|}{Pedestrian} &
  \multicolumn{4}{c}{Cyclist} \\
  \hline
  \multicolumn{1}{c|}{Method} & Easy & Mod & Hard & \multicolumn{1}{|c|}{mAP} & Easy & Mod & Hard & \multicolumn{1}{|c|}{mAP} & Easy & Mod & Hard & \multicolumn{1}{|c}{mAP}\\
  \specialrule{.125em}{0em}{0em}

 \multicolumn{1}{c|}{MVX-Net \cite{MVX-Net}} &
     85.5 & 73.3 & 67.4 & \multicolumn{1}{|c|}{75.4} &
    -- & -- & -- & \multicolumn{1}{|c|}{--} & 
    -- & -- & -- & \multicolumn{1}{|c}{--} \\ 

\multicolumn{1}{c|}{MVX-Net + MoCa \cite{MoCa}} &
    87.9 & 77.6 & 76.0 & \multicolumn{1}{|c|}{80.5} & 
    \textbf{68.6} & 61.9 & 54.7 & \multicolumn{1}{|c|}{61.7} & 
    86.0 & 71.2 & \textbf{65.0} & \multicolumn{1}{|c}{74.1} \\ 

\multicolumn{1}{c|}{MVX-Net + Context \cite{Context-guided}} &
    87.3 & \textbf{79.4} & 74.8 & \multicolumn{1}{|c|}{80.5} & 
    68.2 & \textbf{62.3} & \textbf{57.7} & \multicolumn{1}{|c|}{\textbf{62.7}} & 
    \textbf{86.8} & \textbf{74.6} & 62.8 & \multicolumn{1}{|c}{\textbf{74.7}} \\ 
    \hline

\multicolumn{1}{c|}{MVX-Net + SurfaceAug (Ours)} &
    \textbf{89.2} & 79.1 & \textbf{76.3} & \multicolumn{1}{|c|}{\textbf{81.5}} &
    65.3 & 60.0 & 56.4 & \multicolumn{1}{|c|}{60.6} & 
    82.7 & 63.8 & 60.5 & \multicolumn{1}{|c}{69.0} \\ 
  \end{tabular}
  \label{3d_results_table}
  \caption{3D AP on KITTI validation set. Best results are in bold.}
\end{table*}

\begin{table*}[t]

  \centering
  \begin{tabular}{@{}ccccc|cccc|cccc@{}}

  &
  \multicolumn{4}{|c|}{Car}&
  \multicolumn{4}{|c|}{Pedestrian} &
  \multicolumn{4}{c}{Cyclist} \\
  \hline
  \multicolumn{1}{c|}{Method} & Easy & Mod & Hard & \multicolumn{1}{|c|}{mAP} & Easy & Mod & Hard & \multicolumn{1}{|c|}{mAP} & Easy & Mod & Hard & \multicolumn{1}{|c}{mAP}\\
  \specialrule{.125em}{0em}{0em}

 \multicolumn{1}{c|}{MVX-Net \cite{MVX-Net}} &
     89.5 & 84.9 & 79.0 & \multicolumn{1}{|c|}{84.5} & 
    -- & -- & -- & \multicolumn{1}{|c|}{--} & 
    -- & -- & -- & \multicolumn{1}{|c}{--} \\ 

\multicolumn{1}{c|}{MVX-Net + MoCa \cite{MoCa}} &
    -- & -- & -- & \multicolumn{1}{|c|}{--} & 
    -- & -- & -- & \multicolumn{1}{|c|}{--} & 
    -- & -- & -- & \multicolumn{1}{|c}{--} \\ 


\multicolumn{1}{c|}{MVX-Net + Context \cite{Context-guided}} &
    90.3 & \textbf{88.3} & 84.8 & \multicolumn{1}{|c|}{87.8} & 
    \textbf{77.4} & \textbf{66.9} & \textbf{68.2} & \multicolumn{1}{|c|}{\textbf{70.9}} & 
    \textbf{86.3} & \textbf{82.3} & \textbf{75.1} & \multicolumn{1}{|c}{\textbf{81.2}} \\ 
    \hline

\multicolumn{1}{c|}{MVX-Net + SurfaceAug (Ours)} &
    \textbf{92.7} & 88.2 & \textbf{85.6} & \multicolumn{1}{|c|}{\textbf{88.9}} & 
    71.7 & 66.2 & 62.3 & \multicolumn{1}{|c|}{66.7} & 
    84.1 & 66.1 & 62.2 & \multicolumn{1}{|c}{70.8} \\ 

%
  %
  \end{tabular}
\caption{BEV AP on KITTI validation set. Best results are in bold.}
  \label{bev_results_table}
\end{table*}

\section{Experiments}
\subsubsection{Setup}
For comparison against existing approaches, we partition the KITTI training set into the standard \cite{FrustumConvNet, MV3D, 3DObjectProposals, BiProDet, PV-RCNN, VoxelNet, MVX-Net, MoCa} train/val split of 3,712/3,769 samples, respectively. Cut objects are constructed exclusively from the training set, yielding 2,083 cars, 433 pedestrians, and 65 cyclists. For each frame, we sample from the same distribution of classes used by MoCa \cite{MoCa}: $p(\text{car}) = 0.5$, $p(\text{ped}) = p(\text{cyc}) = 0.25$. However, unlike MoCa, which aims to populate every frame with 12 cars, six pedestrians, and six cyclists, we only paste a maximum of five objects per frame.

We test SurfaceAug by training the MMDetection3D \cite{MMDet3D} implementation of MVX-Net from scratch. Apart from the addition of SurfaceAug, we use the default training configuration. Notably, this includes global point cloud augmentations, horizontal flips, and image rescaling. Global point cloud transformations are inverted as described in MoCa \cite{MoCa} prior to the point fusion module.

\subsubsection{Results}
We evaluate the quality of our detection results on the KITTI validation set using the standard KITTI metrics: average precision (AP) computed across 40 recall positions, with an IoU threshold of 0.7 for cars and 0.5 for pedestrians and cyclists. Notably, our method outperforms the previous state-of-the-art multimodal ground truth sampling algorithm by 1.0 3D mAP and 1.1 BEV mAP on the car class.

We observe that SurfaceAug's performance gain is far greater on cars than on pedestrians and cyclists. We believe there are several reasons for this. First, the car meshes are generally of higher quality than the pedestrian and cyclist meshes because of our point cloud accumulation algorithm, which only works for rigid objects. Second, we select cars for pasting twice as often as pedestrians and cyclists. Finally, like the KITTI dataset it was created from, our cut object database contains many more cars than pedestrians and cyclists, so the pasted cars are more varied.
\section{Conclusion}
We have presented SurfaceAug, a novel ground truth sampling algorithm for image and point cloud data. Unlike previous methods, SurfaceAug fully resamples pasted objects in both modalities and enables out-of-plane object-level transformations, greatly improving data diversity. We show experimentally that SurfaceAug outperforms previous augmentation algorithms for car detection tasks and establishes a new state of the art for multimodal ground truth sampling.
{
    \small
    \bibliographystyle{unsrt}
    \bibliography{main}
}


\end{document}